\documentclass[conference,a4paper]{IEEEtran}
\IEEEoverridecommandlockouts
\usepackage[hidelinks]{hyperref}
\usepackage[cmex10]{amsmath}
\usepackage{amssymb,amsfonts}
\usepackage[ruled,vlined]{algorithm2e}
\usepackage{graphicx}
\graphicspath{{Figures/PDF/}{Figures/PNG/}}

\usepackage{booktabs}
\usepackage{siunitx}
\usepackage[numbers,compress]{natbib}
\usepackage{texnames}
\usepackage{bm,bbm}
\usepackage{orcidlink}
\usepackage{multirow}

\begin{document}

\title{\uppercase{Building-Guided Pseudo-Label Learning for Cross-Modal Building Damage Mapping}
\thanks{Corresponding author: Wei He.}
}

\author{	\IEEEauthorblockN{Jiepan Li}
	\IEEEauthorblockA{\textit{Wuhan University }\\
		430072 Wuhan, China\\
		jiepanli@whu.edu.cn}
	\and
	\IEEEauthorblockN{He Huang}
	\IEEEauthorblockA{\textit{Wuhan University }\\
		430072 Wuhan, China\\
		huang\_{he}@whu.edu.cn}
	\and
	\IEEEauthorblockN{Yu Sheng}
	\IEEEauthorblockA{\textit{Wuhan University }\\
		430072 Wuhan, China\\
		shengyu@whu.edu.cn}
        \and
    \IEEEauthorblockN{Yujun Guo}
	\IEEEauthorblockA{\textit{Wuhan University }\\
		430072 Wuhan, China\\
		yujunguo@whu.edu.cn}
        \and
    \IEEEauthorblockN{Wei He$^*$}
	\IEEEauthorblockA{\textit{Wuhan University }\\
		430072 Wuhan, China\\
		weihe1990@whu.edu.cn}
}

\maketitle
\begin{abstract}
Accurate building damage assessment using bi-temporal multi-modal remote sensing images is essential for effective disaster response and recovery planning. This study proposes a novel Building-Guided Pseudo-Label Learning Framework to address the challenges of mapping building damage from pre-disaster optical and post-disaster SAR images.
First, we train a series of building extraction models using pre-disaster optical images and building labels. To enhance building segmentation, we employ multi-model fusion and test-time augmentation strategies to generate pseudo-probabilities, followed by a low-uncertainty pseudo-label training method for further refinement. Next, a change detection model is trained on bi-temporal cross-modal images and damaged building labels. To improve damage classification accuracy, we introduce a building-guided low-uncertainty pseudo-label refinement strategy, which leverages building priors from the previous step to guide pseudo-label generation for damaged buildings, reducing uncertainty and enhancing reliability.
Experimental results on the 2025 IEEE GRSS Data Fusion Contest dataset demonstrate the effectiveness of our approach, which achieved the highest mIoU score (54.28\%) and secured first place in the competition. The source code will be available at \href{https://github.com/Henryjiepanli/Building-Guided-Pseudo-Label-Learning-for-Cross-Modal-Building-Damage-Mapping}{BGPLL}.
\end{abstract}


\section{Introduction}
Building damage assessment~\cite{zheng2021building} is crucial for post-disaster response and recovery planning, providing essential information for emergency management, resource allocation, and reconstruction efforts. With the increasing availability of remote sensing (RS) data~\cite{huang2024stfdiff}, bi-temporal multi-modal imagery, such as optical and synthetic aperture radar (SAR) images, has become a key resource for damage assessment~\cite{chen2023fourier}. Optical images capture high-resolution visual details under normal lighting conditions~\cite{guo2023blind,li2024overcoming}, while SAR images provide robust structural information regardless of weather and illumination~\cite{huang2024overcoming}. However, integrating these modalities for accurate damage mapping poses significant challenges due to differences in imaging geometry ($e.g.$, perspective and illumination), resolution disparities, inconsistent spatial coverage, and modality-specific noise artifacts such as speckle in SAR, all of which hinder accurate pixel-wise feature alignment.

To advance research in this field, the 2025 IEEE GRSS Data Fusion Contest introduced the "All-Weather Building Damage Mapping” challenge~\cite{chen2025bright}. This competition focuses on assessing building damage using bi-temporal multi-modal images, with a dataset consisting of pre-disaster optical images and post-disaster SAR images at submeter resolution. The dataset is annotated with four classes: background, intact buildings, damaged buildings, and destroyed buildings. A major challenge in this task lies in the substantial differences between optical and SAR imagery, which complicate feature alignment and damage classification. During the evaluation, models are tested on unseen image pairs to generate damage assessments, with performance measured by the mean Intersection over Union (mIoU) metric.

To address these challenges, we propose a Building-Guided Pseudo-Label Learning Framework that leverages building priors to enhance damage classification. Our framework consists of two key stages: building extraction and damaged building change detection. In the first stage, we train a series of building extraction models using pre-disaster optical images and building labels. To further refine building segmentation results, we incorporate multi-model fusion and test-time augmentation strategies to generate pseudo-probabilities for buildings, followed by a low-uncertainty pseudo-label training method. 
We train a change detection model using bi-temporal cross-modal images and damaged building labels based on the reliable building priors from the first stage. 
To improve detection accuracy, we introduce a building-guided low-uncertainty pseudo-label refinement strategy, which leverages building priors from the first stage to guide pseudo-label generation for damaged buildings. This reduces uncertainty in training labels and enhances the model’s ability to distinguish different levels of building damage.

We evaluate our approach on the official test set of the 2025 IEEE GRSS Data Fusion Contest. Experimental results demonstrate the effectiveness of our framework, achieving the highest mIoU score of 54.28\%, ranking first place in the competition. These findings highlight the potential of our method for improving building damage mapping accuracy in real-world disaster scenarios.
\begin{figure*}[t]
\centering
\includegraphics[width=0.9\linewidth]{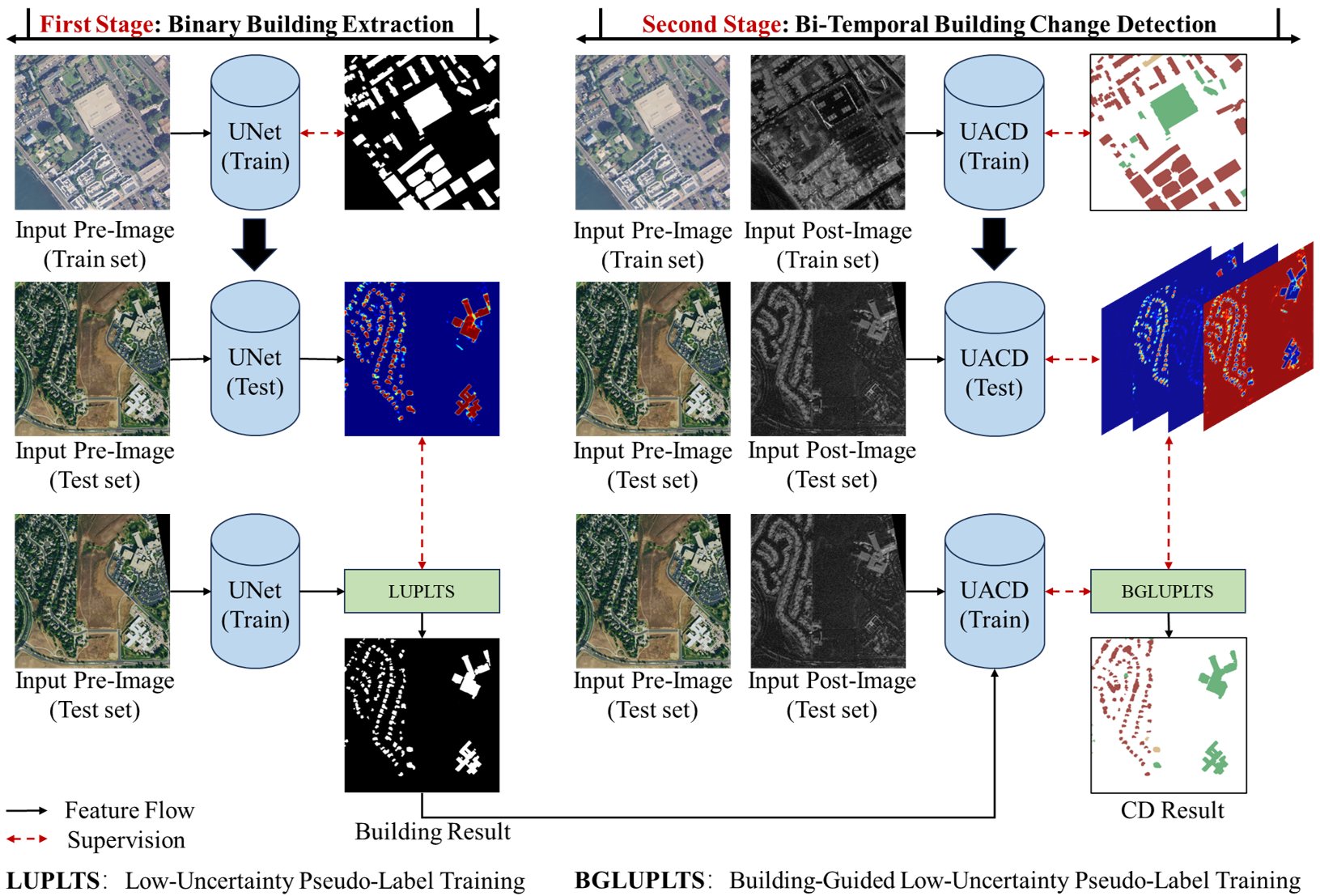}
\vspace{-0.5em}
\caption{Architecture of our proposed Cross-Modal Building Damage Mapping Framework.}
\label{fig:framework}
\vspace{-1.5em}
\end{figure*}
The main contributions of this paper are summarized as follows:
\begin{enumerate}[]
\item We propose a Building-Guided Pseudo-Label Learning Framework that integrates building priors to improve cross-modal building damage mapping.
\item We introduce a low-uncertainty pseudo-label refinement strategy to enhance label reliability and reduce the impact of modal discrepancies.
\item Our framework achieves state-of-the-art performance, ranking first in the 2025 IEEE GRSS Data Fusion Contest, demonstrating its effectiveness in real-world disaster assessment.
\end{enumerate}

The remainder of this paper is organized as follows. Section 2 details our proposed framework, including building segmentation, damaged building change detection, and pseudo-label refinement. Section 3  and Section 4 present experimental results and the corresponding analysis. Finally, Section 5 concludes the paper and discusses future research directions.
\section{Methodology}

\subsection{Overview}

As illustrated in Fig.~\ref{fig:framework}, our building-guided pseudo-label learning framework comprises two main stages. In the first stage, we train multiple building extraction models using pre-disaster optical images and their corresponding building annotations. To improve the quality of segmentation results, we employ a multi-model fusion strategy and test-time augmentation to generate pseudo-probability maps, which are subsequently refined using a low-uncertainty guided training strategy.

In the second stage, we train a change detection model (UACD~\cite{UABCD}) using bi-temporal and cross-modal images, along with damaged building annotations. To further enhance detection performance, we introduce a building-guided pseudo-label refinement approach, leveraging building priors from the first stage. This method helps reduce the uncertainty of the label and strengthens the model's ability to accurately differentiate between the varying levels of damage to the building.

\subsection{Binary Building Extraction}
This stage involves three major components: (1) Building Extraction Model Training, (2) Pseudo-Label Generation, and (3) Low-Uncertainty Pseudo-Label Training. Initially, we train multiple segmentation models using labeled pre-disaster optical images. Next, ensemble learning and test-time augmentation are applied to produce reliable pseudo-probability maps. Finally, a selective training scheme guided by uncertainty is adopted to refine the predictions. Each step is detailed below.

\subsubsection{Building Extraction Model Training}
Following previous works~\cite{UANet,Comprehensive_farmland}, modeling long-range dependencies is crucial for accurate segmentation in RS imagery. To this end, we adopt three variants of the Pyramid Vision Transformer~\cite{pvtv2} (PVT-v2-b2, PVT-v2-b3, and PVT-v2-b4) as encoders, each coupled with a Feature pyramid network(FPN~\cite{fpn})-style decoder for dense prediction.

The official dataset is partitioned into 80\% for training and 20\% for validation. Only pre-disaster optical images are used as input, with binary labels supervising the training via standard cross-entropy loss. We retain the best-performing model from each configuration based on validation performance, resulting in three models: $M_{1}$, $M_{2}$, and $M_{3}$.

\subsubsection{Pseudo-Label Generation}
Using the trained models ($M_{1}$, $M_{2}$, and $M_{3}$), we generate pseudo-labels for pre-disaster images in the test set through an ensemble strategy augmented with test-time data augmentation. Given an input image $I_{1}$, we first apply horizontal flipping to obtain its augmented version $I_{2}$. 
The three models process both $I_{1}$ and $I_{2}$, and the outputs are averaged separately. 
The final pseudo-probability map is calculated by combining the original and flipped predictions as follows: 
\begin{equation} 
\begin{split} 
P_{1} &= \frac{M_1(I_{1}) + M_2(I_{1}) + M_3(I_{1})}{3}, \\ 
P_{2} &= \frac{M_1(I_{2}) + M_2(I_{2}) + M_3(I_{2})}{3}, \\ 
P_{f} &= \frac{P_{1} + De(P_{2})}{2}, 
\end{split} 
\end{equation} 
where $De(\cdot)$ denotes the inverse horizontal flip operation to align $P_{2}$ with the original spatial orientation.

\subsubsection{Low-Uncertainty Pseudo-Label Training}
To mitigate the influence of noisy pseudo-labels, we design a low-uncertainty guided training strategy. From the pseudo-probability map $P_{f}$, we construct two types of supervision signals: a hard label $G_{h}$ and a soft label $G_{s}$, defined as follows: \begin{equation}
\begin{aligned}
G_{h}(x) &= 
\begin{cases}
1, & \text{if } P_{f}(x) \geq 0.5, \\
0, & \text{otherwise},
\end{cases} \\
G_{s}(x) &= \left[ G_{h}(x),\; P_{f}(x) \right],
\end{aligned}
\end{equation}
where $x$ represents a spatial location, and $[\cdot]$ denotes channel-wise concatenation.

During training, the model prediction $O$ for an input $I$ is used to compute a pixel-wise entropy map, representing prediction uncertainty. This entropy is normalized to $E_{n} \in [0,1]$. Pixels with $E_{n} < 0.3$ are considered reliable and supervised using the hard label $G_{h}$ via cross-entropy loss. For uncertain pixels, we employ a self-training scheme by aligning the model outputs with $G_{s}$ through Kullback–Leibler (KL) divergence.


\subsection{Bi-Temporal Building Change Detection}
This stage follows a similar structure to binary building extraction, consisting of three key steps: (1) Change Detection Model Training, (2) Pseudo-Label Generation, and (3) Building-Guided Low-Uncertainty Pseudo-Label Training.

\subsubsection{Change Detection Model Training}
We adopt the UACD~\cite{UABCD} architecture, removing its uncertainty modeling components to reduce computational overhead while retaining the core bi-temporal fusion mechanism. The model is trained on the image pairs from the official training split and evaluated on the Phase 1 validation set. The top five models are selected for ensemble construction to generate pseudo-labels for the test set in the next stage.
\begin{figure*}[t]
    \centering

    \begin{minipage}[t]{2.9cm}
        \centering
        \includegraphics[width=2.9cm]{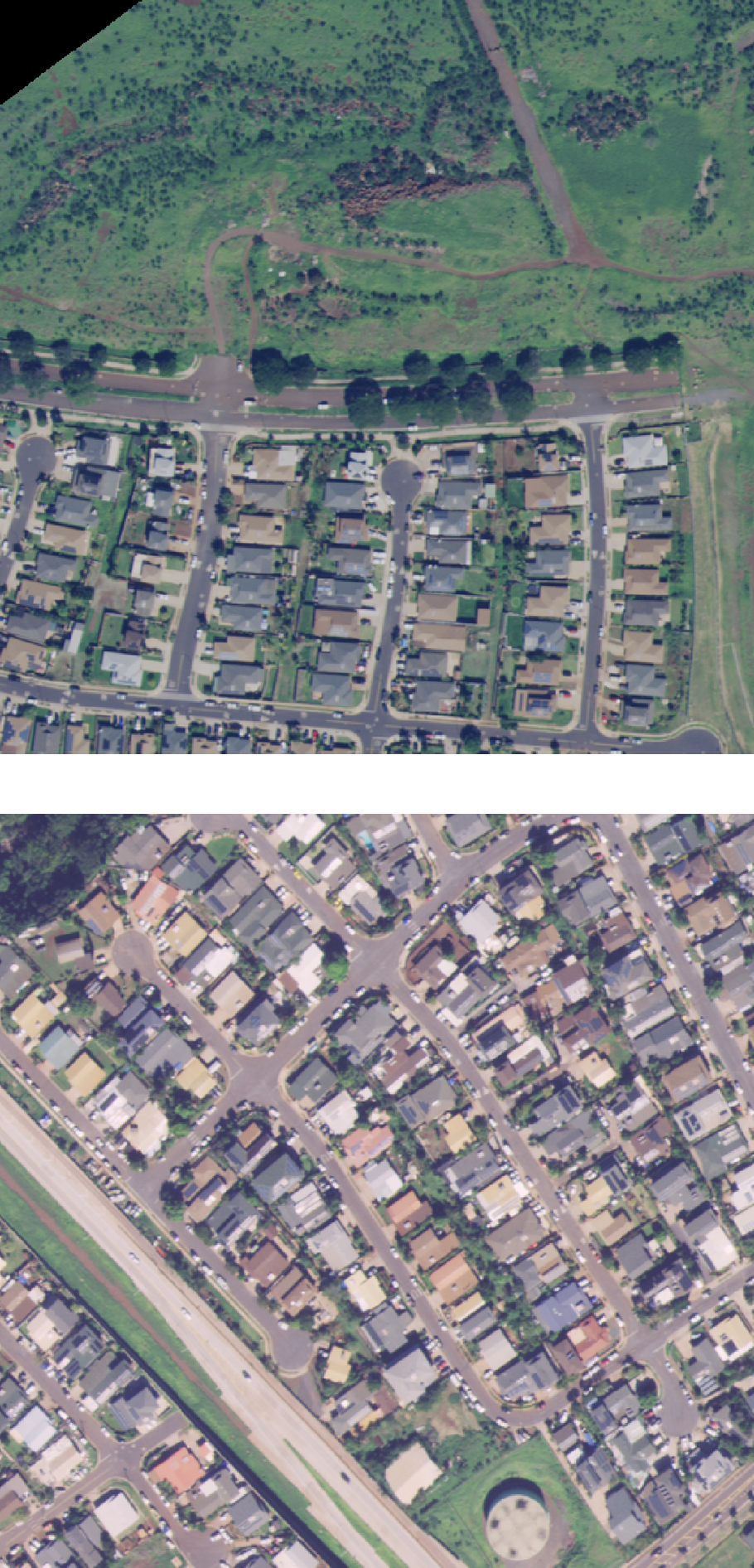}
        
        {\footnotesize (a) Pre Disaster}
    \end{minipage}
    \hfill
    \begin{minipage}[t]{2.9cm}
        \centering
        \includegraphics[width=2.9cm]{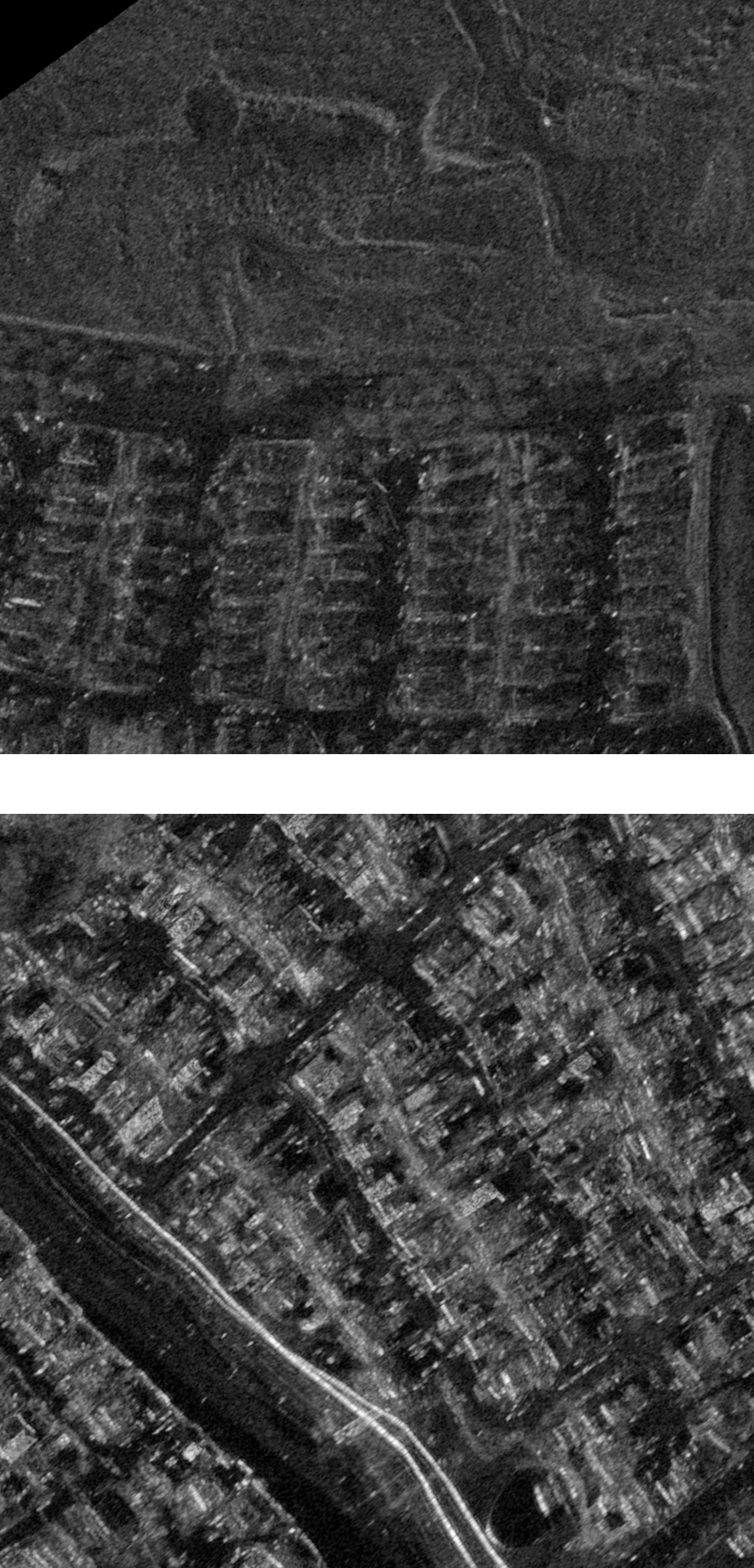}
        
        {\footnotesize (b) Post Disaster}
    \end{minipage}
    \hfill
    \begin{minipage}[t]{2.9cm}
        \centering
        \includegraphics[width=2.9cm]{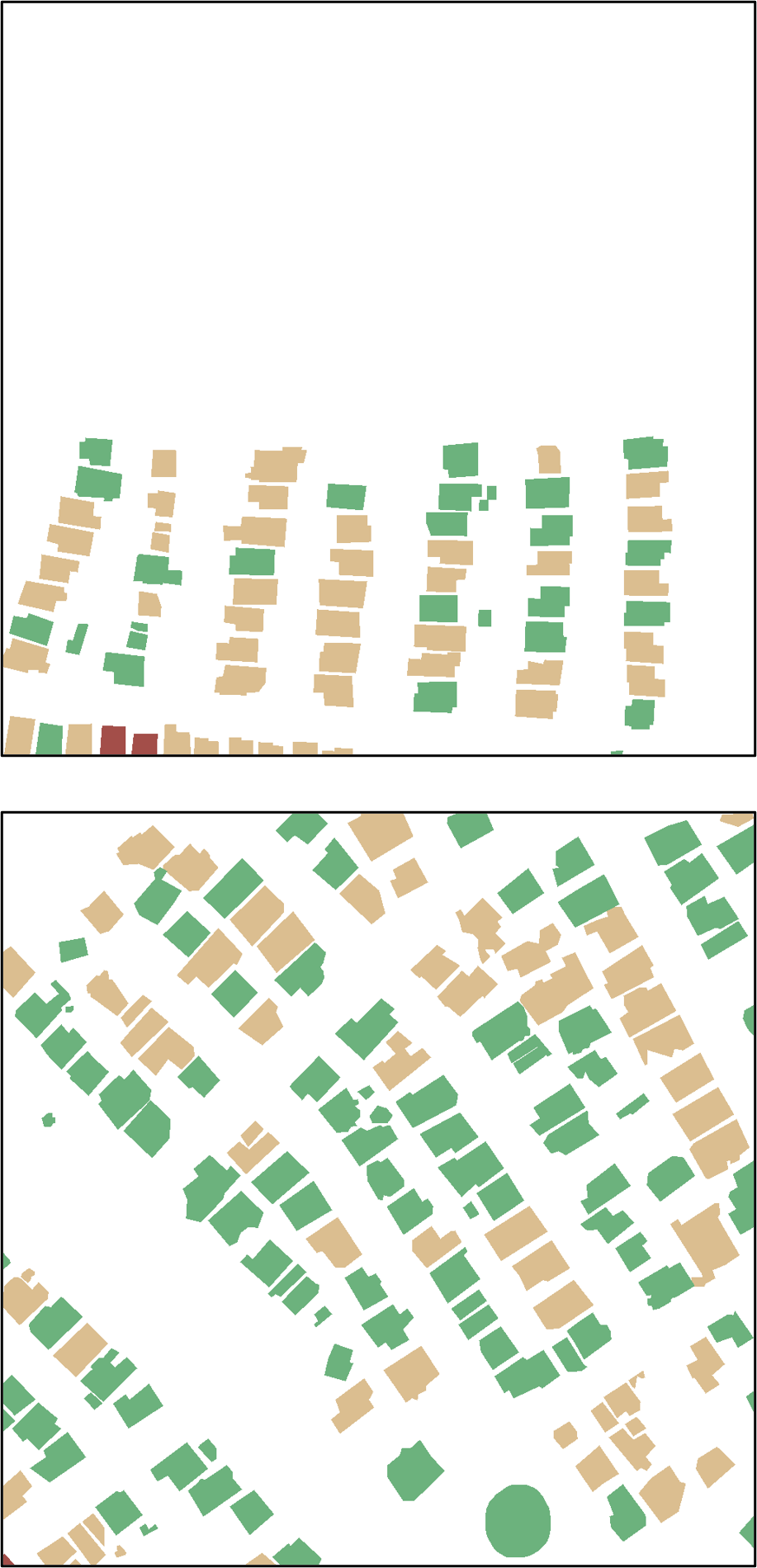}
        {\footnotesize (c) Ground Truth}
    \end{minipage}
    \hfill
    \begin{minipage}[t]{2.9cm}
        \centering
        \includegraphics[width=2.9cm]{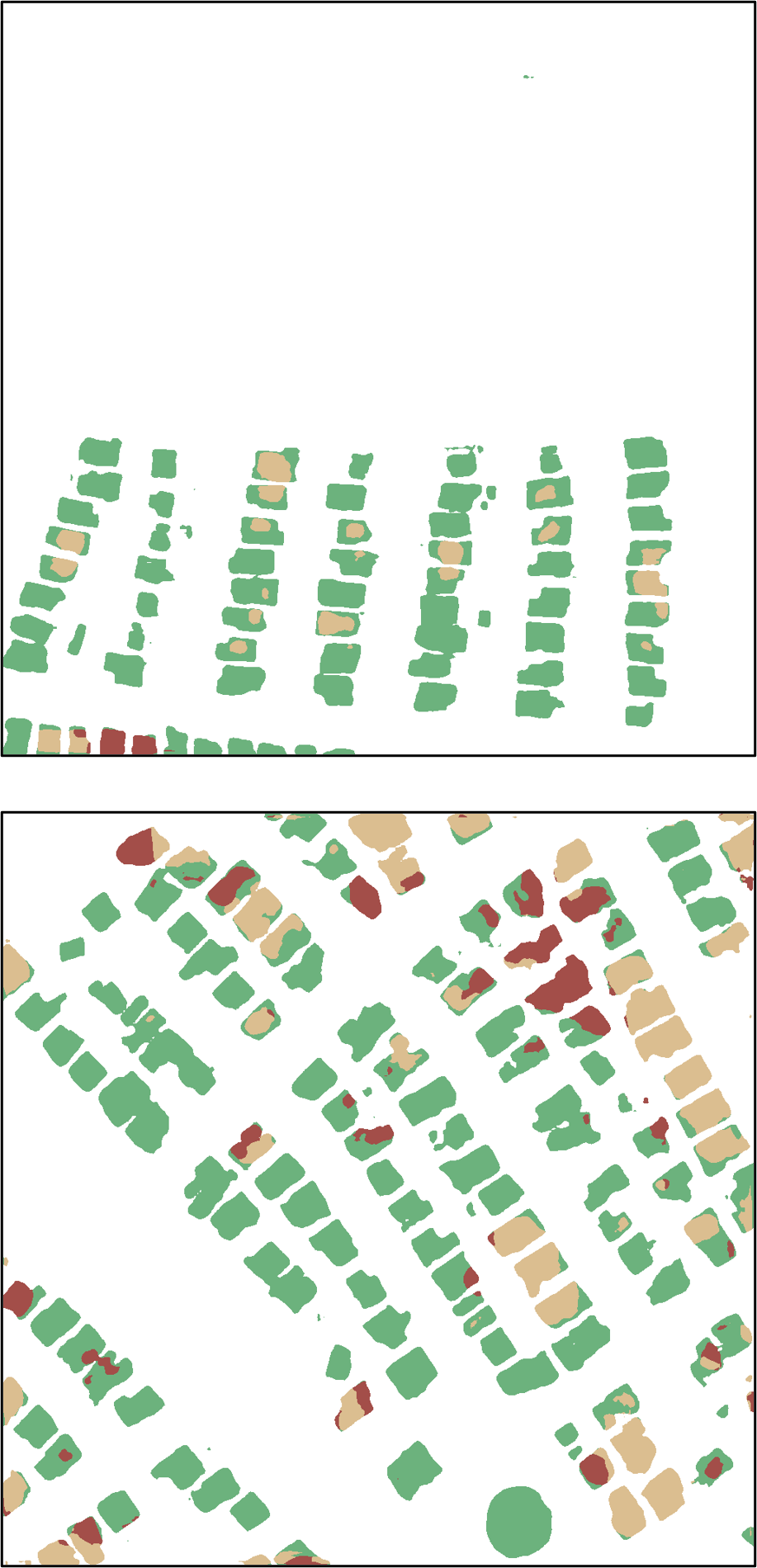}
        
        {\footnotesize (d) ChangeMamba}
    \end{minipage}
    \hfill
    \begin{minipage}[t]{2.9cm}
        \centering
        \includegraphics[width=2.9cm]{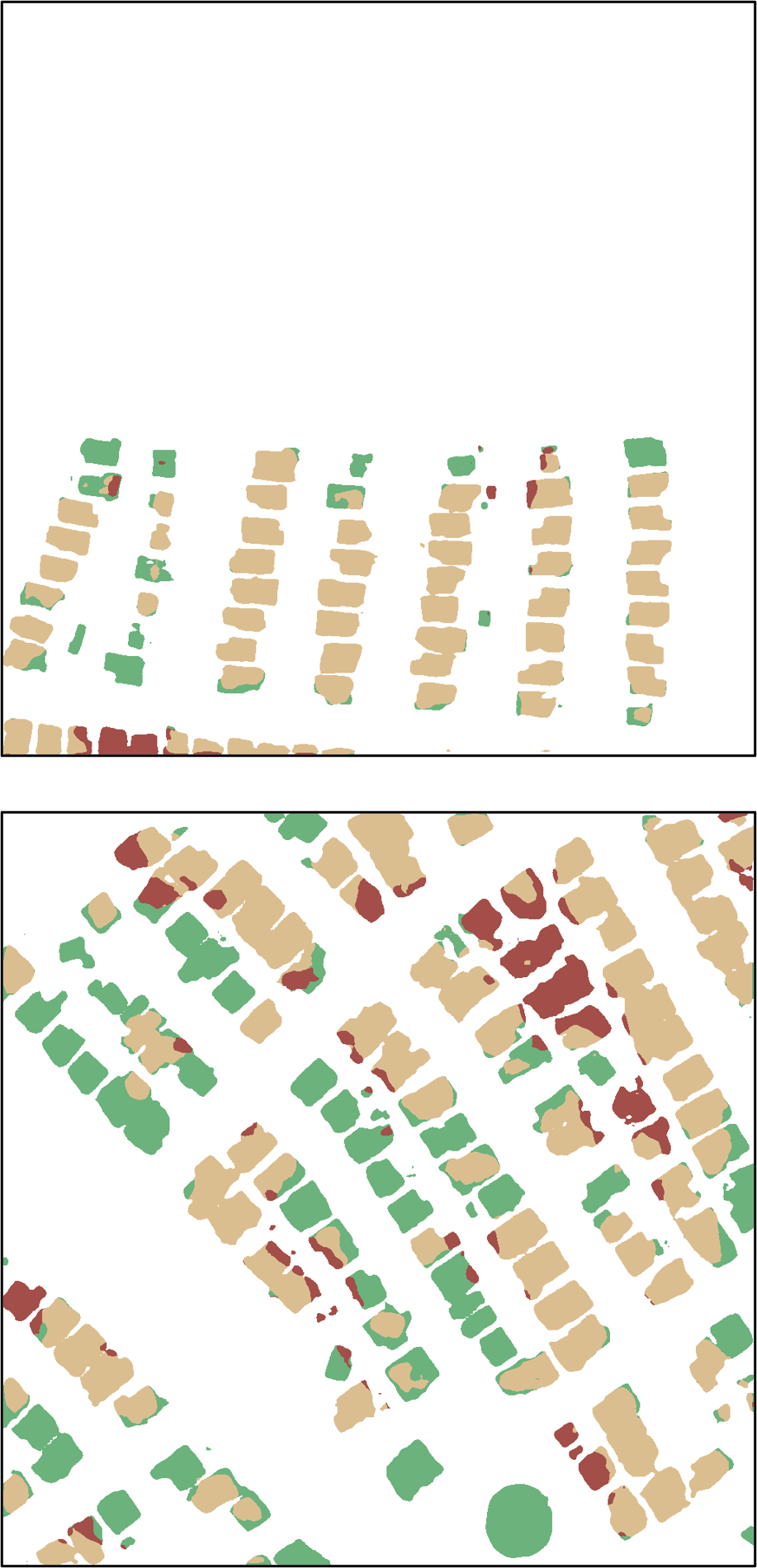}      
        {\footnotesize (e) UACD}
    \end{minipage}
    \hfill
    \begin{minipage}[t]{2.9cm}
        \centering
        \includegraphics[width=2.9cm]{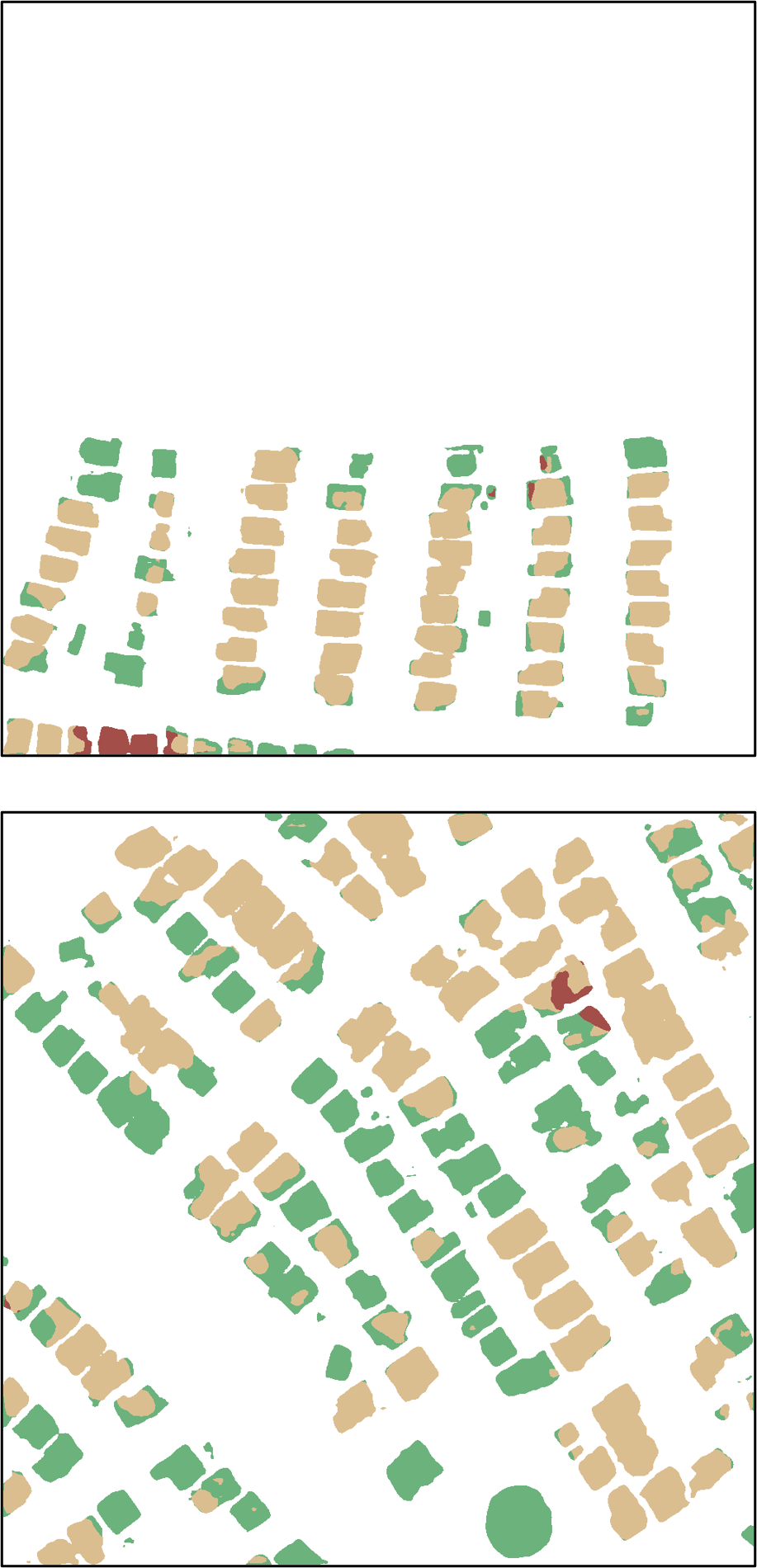}
        {\footnotesize (f) Ours}
    \end{minipage}
    \caption{Visual comparison between our proposed framework and the compared methods. Notably, white, green, yellow, and red denote the background, intact buildings, damaged buildings, and destroyed buildings, respectively.}
    \label{fig_vis}
    \vspace{-1.5em}
\end{figure*}
\subsubsection{Pseudo-Label Generation}
We generate pseudo-labels $P_{cd}$ for the bi-temporal image pairs in the test set using the ensemble of top-performing models combined with test-time augmentation. The procedure mirrors that of the binary building extraction stage. Each pseudo-label $P_{cd}$ is a four-channel probability map, where each channel corresponds to the predicted probability of a specific damage category.

\subsubsection{Building-Guided Low-Uncertainty Pseudo-Label Training}
To address noise in the pseudo-labels, we propose a building-guided low-uncertainty training strategy. Given the pseudo-probability map $P_{cd}$, we define a hard label $G_{h}^{cd}$ and a soft label $G_{s}^{cd}$ as: 
\begin{equation} 
\begin{aligned} 
G_{h}^{cd}(x) &= \arg\max_{c} P_{cd}^{(c)}(x), \\
G_{s}^{cd}(x) &= \left[ G_{h}^{cd}(x), P_{cd}(x) \right], \end{aligned} 
\end{equation} 
where $x$ denotes a pixel location and $[\cdot]$ indicates channel-wise concatenation.

During training, bi-temporal features extracted by the encoder (PVT-v2-b2) are modulated using the binary building map obtained in the first stage. This guidance helps enhance building-related features and preserves structural consistency. Given a bi-temporal input pair $(I_{pre}, I_{post})$, the model generates an output $O^{cd}$, from which a pixel-wise entropy map is derived to estimate prediction uncertainty. The normalized entropy $E_{n}^{cd} \in [0,1]$ is used to distinguish between reliable and uncertain regions.

Pixels with $E_{n}^{cd} < 0.3$ are treated as confident and supervised using cross-entropy loss with hard labels $G_{h}^{cd}$. In contrast, for high-uncertainty pixels, we apply a self-regularization scheme based on KL divergence with soft labels $G_{s}^{cd}$.


\section{Experiments}
\subsection{Experimental Settings}
All experiments were conducted using PyTorch 1.8.1 (CUDA 11.1) on an NVIDIA RTX 3090 GPU (24 GB). Models were trained for 100 epochs with the AdamW~\cite{AdamW} optimizer (\(\beta_1 = 0.9\), \(\beta_2 = 0.999\)) and a cosine annealing schedule starting at \(10^{-4}\). The loss function combines cross-entropy and KL divergence with a fixed $2:1$ weight ratio, supervising low- and high-uncertainty regions, respectively. All images were kept at their original \(1024 \times 1024\) resolution.
For building extraction, we ensembled PVT-v2-b2 and PVT-v2-b3 models trained with our Low-Uncertainty Pseudo-Label framework, and applied horizontal flip-based test-time augmentation (TTA). For change detection, we used a single PVT-v2-b2 model trained with the Building-Guided strategy, also evaluated with horizontal flip TTA.

\subsection{Visual Comparison}
As illustrated in Fig.~\ref{fig_vis}, we present two representative examples to demonstrate the superiority of our proposed method. It is evident that the compared change detection method, ChangeMamba~\cite{ChangeMamba}, fails to identify many damaged buildings. Although UACD achieves more accurate detection, it still misclassifies some damaged buildings as destroyed ones. In contrast, our building-guided pseudo-label learning strategy effectively leverages building information and utilizes low-uncertainty pseudo labels to guide the model in learning richer damage-aware representations.
\subsection{Quantitative Comparison}
\begin{table}[t]
\small
\setlength\tabcolsep{2pt}
\caption{Quantitative results on the Development Phase and Test Phase.}
\vspace{-0.8em}
\label{tab:1}
\centering
\begin{tabular}{c|c|c|c|c|c}
\hline
\multirow{2}{*}{Method} & \multicolumn{4}{c|}{Development Phase IoU (\%)} &\multirow{2}{*}{mIoU} \\ 
\cline{2-5}
 & Background & Intact & Damaged & Destroyed  \\ 
\hline
CGNet             & 96.722 & 79.519 & 31.000 & 70.022 & 69.316 \\
ChangeMamba       & 96.737 & 80.022 & 38.181 & 71.464 & 71.601 \\
UACD              & 96.756 & 80.076 & 39.575 & 73.148 & 72.389 \\
Ours (w/o bg)     & 96.811 & 80.700 & 44.618 & 74.105 & 74.105 \\
Ours (with bg)    & \textbf{96.845} & \textbf{80.958} & \textbf{45.321} & \textbf{74.637} & \textbf{74.440} \\
\hline
\multirow{2}{*}{Method} & \multicolumn{4}{c|}{Test Phase IoU (\%)} &\multirow{2}{*}{mIoU} \\ 
\cline{2-5}
 & Background & Intact & Damaged & Destroyed \\  
\hline
Ours (with bg)    & \textbf{97.995} & \textbf{67.304} & \textbf{9.748} & \textbf{42.086} & \textbf{54.283} \\
\hline
\end{tabular}
{bg denotes the use of the proposed building-guided strategy.}
\vspace{-1.5em}
\end{table}
As shown in Tab.~\ref{tab:1}, we compare the performance of our proposed framework with several state-of-the-art (SOTA) change detection methods, including Change Guiding Network (CGNet\cite{CGNet}), ChangeMamba~\cite{ChangeMamba}, and the Uncertainty-aware Change Detection Framework (UACD~\cite{UABCD}). We also evaluate the effectiveness of our building-guided pseudo-label learning strategy by comparing the results with and without its incorporation. The results clearly demonstrate that our method significantly outperforms the competing approaches in terms of Intersection over Union (IoU) across all four categories.

In the development phase, where validation scenes are similar to training data, all methods achieve relatively strong performance in mapping damaged buildings. However, during testing, with differing disaster scenes, most methods see a performance drop. Our framework, however, maintains high accuracy and robustness, helping our team achieve first place in the 2025 IEEE GRSS Data Fusion Contest.
\vspace{-0.5em}

\section{Discussion}
\begin{figure}[]
\centering
\includegraphics[width=1.05\linewidth]{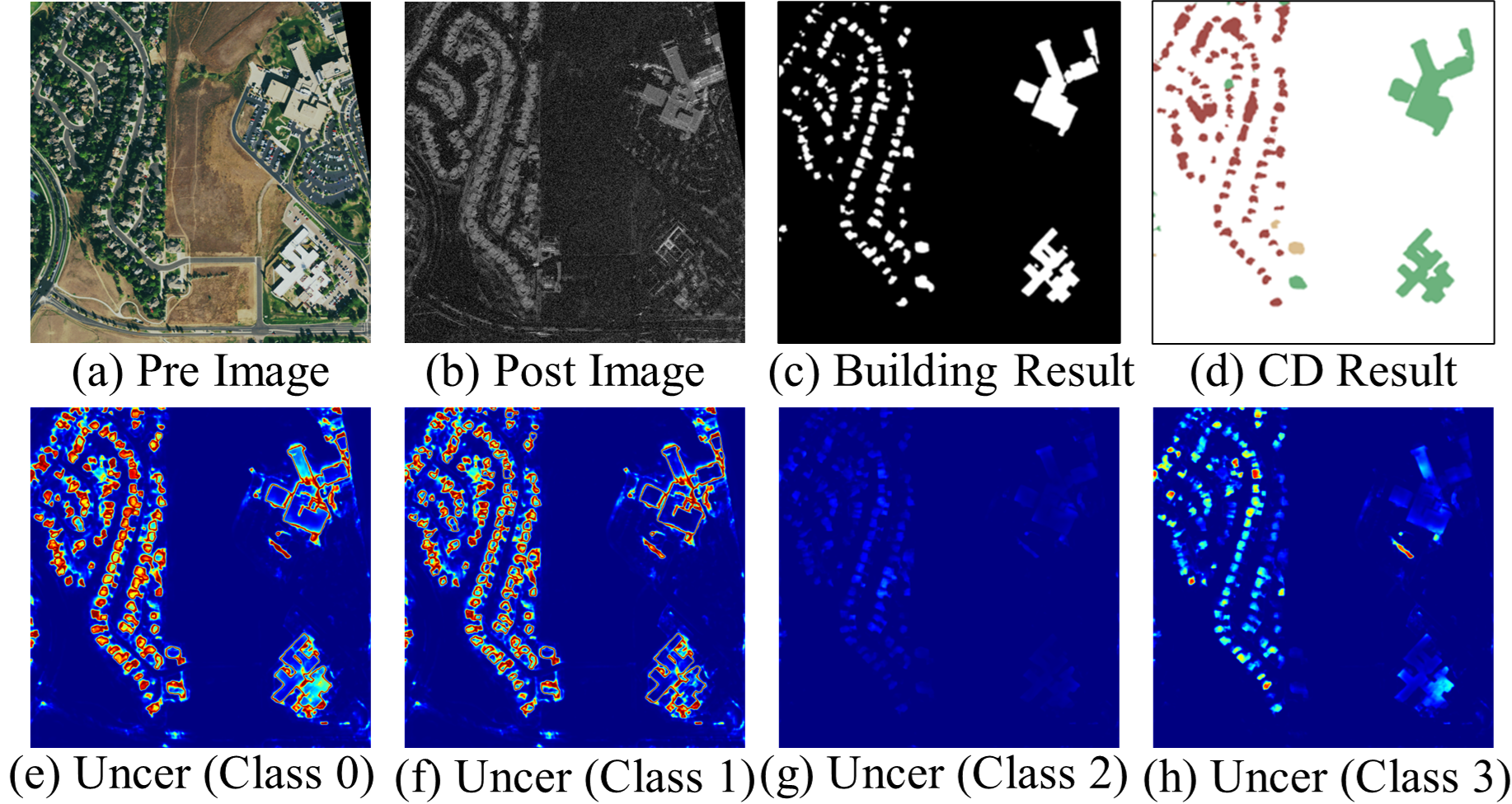}
\vspace{-1.5em}
\caption{Visualization of pseudo-label uncertainty (uncer).}
\label{fig:discussion}
\vspace{-2.0em}
\end{figure}

To evaluate the effectiveness of the low-uncertainty pseudo-label training strategy, we visualize the uncertainty associated with the pseudo-labels. As shown in Fig.~\ref{fig:discussion}, the pseudo-labels exhibit considerable uncertainty, particularly along building boundaries. However, our strategy selectively leverages the reliable, low-uncertainty predictions to guide the model in learning more accurate and damage-aware representations.

Distinguishing partially damaged buildings is difficult due to their blend of intact features and damage, along with challenges like segregated datasets, varying damage patterns, and resolution differences. 
As seen in the first phase, without isolated validation and test scenarios, the ``damaged" class had significantly lower IoU, reflecting the model's struggle in diverse disaster environments. Our uncertainty-guided approach, incorporating building priors, showed slight improvement.
However, to further enhance detection, generative methods to synthesize more partially damaged samples could provide the model with better training data for realistic damage patterns.

\vspace{-0.5em}
\section{Conclusion}
In this paper, we proposed a building-guided low-uncertainty pseudo-label training framework for cross-modal bi-temporal damaged building mapping. Our method achieved first place in Track 2 of the 2025 IEEE GRSS Data Fusion Contest. In future work, we plan to further explore strategies for improving the generalization ability of damaged building mapping across diverse disaster scenes.

\small
\bibliographystyle{IEEEtranN}


\end{document}